\documentclass[sigconf]{acmart}
\renewcommand\footnotetextcopyrightpermission[1]{}
\usepackage{setspace, subfigure}
\usepackage{algorithm}
\usepackage{algorithmic}
\usepackage{multirow}
\usepackage{booktabs}
\usepackage{colortbl}
\definecolor{mygray}{gray}{.9}
\definecolor{mypink}{rgb}{.99,.91,.95}
\definecolor{mycyan}{cmyk}{.3,0,0,0}

\AtBeginDocument{%
  \providecommand\BibTeX{{%
    \normalfont B\kern-0.5em{\scshape i\kern-0.25em b}\kern-0.8em\TeX}}}

\setcopyright{acmcopyright}
\copyrightyear{2018}
\acmYear{2018}
\acmDOI{10.1145/1122445.1122456}

\begin{document}

\title{Orderly Dual-Teacher Knowledge Distillation for Lightweight Human Pose Estimation}
\author{Zhong-Qiu Zhao$^{1}$, Yao Gao$^{1}$, Yuchen Ge$^{1}$, Weidong Tian$^{1}$}
\affiliation{$^{1}$Hefei University of Technology}
\email{{z.zhao, wdtian}@hfut.edu.cn, {2018110922, geyuchen}@mail.hfut.edu.cn}



\begin{abstract}
  Although deep convolution neural networks (DCNN) have achieved excellent performance in human pose estimation, these networks often have a large number of parameters and computations, leading to the slow inference speed. For this issue, an effective solution is knowledge distillation, which transfers knowledge from a large pre-trained network (teacher) to a small network (student). However, there are some defects in the existing approaches: (I) Only a single teacher is adopted, neglecting the potential that a student can learn from multiple teachers. (II) The human segmentation mask can be regarded as additional prior information to restrict the location of keypoints, which is never utilized. (III) A student with a small number of parameters cannot fully imitate heatmaps provided by datasets and teachers. (IV) There exists noise in heatmaps generated by teachers, which causes model degradation. To overcome these defects, we propose an orderly dual-teacher knowledge distillation (ODKD) framework, which consists of two teachers with different capabilities. 
  Specifically, the weaker one (primary teacher, PT) is used to teach keypoints information, the stronger one (senior teacher, ST) is utilized to transfer segmentation and keypoints information by adding the human segmentation mask.
  Taking dual-teacher together, an orderly learning strategy is proposed to promote knowledge absorbability. Moreover, we employ a binarization operation which further improves the learning ability of the student and reduces noise in heatmaps. Experimental results on COCO and OCHuman keypoints datasets show that our proposed ODKD can improve the performance of different lightweight models by a large margin, and HRNet-W16 equipped with ODKD achieves state-of-the-art performance for lightweight human pose estimation.
\end{abstract}




\maketitle

\section{Introduction}
Human pose estimation aims at locating the human keypoints in the input images. As a fundamental computer vision task, it has been applied in many areas such as human action recognition, virtual reality and smart surveillance.

With the development of DCNN~\cite{newell2016stacked, chen2018cascaded, xiao2018simple, sun2019deep,li2019rethinking,cai2020learning}, human pose estimation has achieved significant improvements. On the challenging COCO benchmark~\cite{lin2014microsoft}, these networks consistently achieve top accuracy. However, the performance improvements always come with the cost of increasing the amount of parameters and computations, which leads to poor practicabilities on embedded devices such as smart phones and robots, like high memory requirements and slow inference speed.
\begin{figure}[t]
	\setlength{\belowcaptionskip}{-0.5cm}
	\begin{minipage}[b]{1.0\linewidth}
		\centering
		\centerline{\includegraphics[scale=0.03]{7.pdf}}
	\end{minipage}
	\caption{The process of introducing prior information.}
	\label{fig:motivation}
\end{figure}
Li \emph{et al.}~\cite{li2020human} successfully constructed an in-home lower body rehabilitation system based on lightweight HRNet which makes the lightweight models receive more attention. Therefore, how to make a trade-off between performance and efficiency has become a crucial problem.


There have been some attempts to address the above problem. On the basis of the large models~\cite{cao2017realtime, xiao2018simple, sun2019deep}, recent works~\cite{osokin2018real, zhang2019simple, li2020human} employed few stages or small backbones to achieve model compression. Although these models achieve a faster inference speed, their performance has a dramatic drop. Among the model compression approaches, knowledge distillation shows great superiority, which is to transfer knowledge of a large teacher model to a small student model. And a current work~\cite{zhang2019fast} has employed the student-teacher strategy to improve the performance and speed of the a student model. However, we observe that this method suffers from several problems: (I) Only a single teacher is used in the overall process, neglecting that multiple teachers can provide more privileged information for a student. (II) As a detection task, human pose estimation consists of two subtasks: the classification task (identifying the keypoints) and the regression task (locating the keypoints). It is difficult for large models to achieve high detection accuracy and more difficult for small lightweight models. Meanwhile, lightweight models are obtained by using some operations such as fewer stages and channels, which further causes inefficient heatmaps learning. (III) Ideally, in heatmaps generated by a human pose estimation model, there is only a peak with maximum response corresponding to the position of keypoints. However, multiple peaks usually appear in heatmaps, as pointed out in DARK~\cite{zhang2020distribution}. Among them, only a peak is what we need to approximate and imitate. Redundant peaks can be deemed noise, which provides the wrong learning signal for a student model and causes the degeneration of a student model.
Moreover, the existing human pose estimation models suffer from a common issue: in crowded scenarios, keypoints of one person are easily located on the body of another person, which severely affects the final performance. As available and extra prior information, as shown in Figure~\ref{fig:motivation}, the human segmentation mask can provide valuable context cues to help restrict the position of keypoints, however, which is ignored.



To solve the above problems, we propose a new learning framework called orderly dual-teacher knowledge distillation (ODKD), which introduces two teachers with different capabilities. Specifically, the weaker one (primary teacher, PT) transfers keypoints information to a student, the stronger one (senior teacher, ST) teaches segmentation and keypoints information. These two teachers have slightly different inputs and network structures. Besides, as shown in TAKD~\cite{mirzadeh2020improved}, a student network performance degrades when the gap between a student and teacher is large. Considering that a similar situation also exists in our method, so we adopt the same strategy in TAKD where PT serves as a teacher assistant to bridge the gap between a student and ST. 


Dual-teacher is employed in our proposed ODKD framework, a natural question to ask is: could we adopt more teachers to provide a student with more learning signals? In this paper, we mainly consider the following two aspects. Our proposed dual-teacher has different assignments where one teaches keypoints information, the other transfers segmentation and keypoints information. Adding additional teachers does not introduce new information. Besides, taking a dual-teacher is a compromise between performance and efficiency. The more teachers are, the better the performance may be but the longer the training time is. And we focus on exploring the knowledge distillation framework, rather than designing lightweight network structures. Therefore, we utilize two teachers in the final framework.

In a real-world scenario, when a small student network is asked to learn from two large teacher networks with different capabilities, a straightforward method is to learn from dual-teacher simultaneously. However, when facing two learning signals, the student is confused about which teacher it should learn from, and yielding a suboptimal result. From the perspective of human cognition, a more realistic and reasonable solution is multi-step learning. In theory, multi-step learning can stimulate more efficient learning of a student network. To this end, we propose an orderly learning strategy where the student learns from ST and PT successively so that knowledge from two aspects can be fully absorbed. To further improve the learning efficiency of a student network and reduce noise in heatmaps generated by teachers, we employ a binarization operation in~\cite{chen2018multi} by which the student network only needs to classify each pixel in heatmaps as 0 or 1. And an appropriate binarization threshold can erase extra peaks, which avoids ineffective learning and model degeneration.

The purpose of adopting binarization operation in this paper is different from~\cite{chen2018multi}. In~\cite{chen2018multi}, the binarization operation is utilized to convert the task so that Focal Loss can be used to address the class imbalance problem in human pose estimation. However, we employ a binarization operation to simplify the learning task and reduce noise.


To demonstrate the effectiveness and extensibility of our proposed ODKD framework, we conduct a series of experiments on two keypoints datasets, COCO~\cite{lin2014microsoft} and OCHuman~\cite{zhang2019pose2seg}. Experimental results show that ODKD can promote the existing lightweight human pose estimation models by a large margin.

The contributions of this paper are summarized as follows:
\begin{itemize}
	\setlength{\itemsep}{0pt}
	\setlength{\parsep}{0pt}
	\setlength{\parskip}{0pt}
	\item In contrast to the existing works focusing on improving the performance of human pose estimation models, we pay more attention to the model efficiency. To this end, we propose an orderly dual-teacher knowledge distillation framework (ODKD) for human pose estimation, which integrates an orderly dual-teacher and the human segmentation mask. The proposed ODKD framework serves as a model-agnostic approach and can be applied to most of lightweight human pose estimation models.
	\item We adopt a binarization operation to convert the regression task to the classification task, and the task after binarization is simpler than the original task.
	\item We verify the effectiveness and extensibility of ODKD on different benchmark datasets, COCO and OCHuman with different baseline models. And HRNet-W16 equipped with ODKD achieves state-of-the-art performance for lightweight human pose estimation.
\end{itemize}

\section{Related Work}

\subsection{Lightweight Human Pose Estimation}
There have been some works to compress the human pose estimation models. Based on OpenPose~\cite{cao2017realtime}, Daniil \emph{et al.}~\cite{osokin2018real} proposed a lightweight OpenPose network where the heavy computational backbone VGG~\cite{simonyan2014very} was replaced by the simple and effective MobileNet~\cite{howard2017mobilenets}, and $7\times7$ convolutions were replaced by lots of $3\times3$ convolutions. Although its inference speed becomes faster, there is a significant performance gap between the model and the current mainstream models. Umer \emph{et al.}~\cite{rafi2016efficient} constructed an efficient convolutional network to accelerate inference without conducting quantitative experiments on model efficiency. Bulat \emph{et al.}~\cite{bulat2017binarized} employed the neural network binarization to achieve model compression. Zhang \emph{et al.}~\cite{zhang2019simple} proposed a lightweight pose network (LPN), which was equipped with the depthwise separable convolution and iterative training strategy. However, 
this iterative training strategy takes more than triple the training time of the original training strategy. Li \emph{et al.}~\cite{li2020human} proposed a lightweight HRNet which integrated the attention mechanism with Efficient Spatial Pyramid (ESP)~\cite{mehta2018espnet}.

\subsection{Knowledge Distillation}
Knowledge distillation has been successfully applied to many computer vision tasks, such as image classification~\cite{komodakis2017paying, yim2017gift, wang2018kdgan, mirzadeh2020improved, son2020densely}, object detection~\cite{chen2017learning, dai2021general}, pose estimation~\cite{zhang2019fast, zhao2018through}, etc. Chen \emph{et al.}~\cite{chen2017learning} employed knowledge distillation to learn compact object detection networks and proposed several loss functions to improve the efficiency of knowledge transfer. Dai \emph{et al.}~\cite{dai2021general} proposed a general instance distillation for object detection where feature-based, relation-based and response-based knowledge was considered. Zhao \emph{et al.}~\cite{zhao2018through} employed a knowledge framework to solve the occlusion problem in human pose estimation. Zhang \emph{et al.}~\cite{zhang2019fast} expanded the lightweight human pose estimation network by introducing knowledge distillation, but the approach only employed a single teacher, neglecting that multiple teachers can provide more valuable information. Mirzadeh \emph{et al.}~\cite{mirzadeh2020improved} introduced 
a distillation framework called Teacher Assistant Knowledge Distillation (TAKD), where an intermediate-sized network (teacher assistant) was adopted to bridge the gap between the student and the teacher. Song \emph{et al.}~\cite{son2020densely} proposed a densely guided knowledge distillation (DGKD) to eliminate the error avalanche problem in TAKD. The main difference between our method and TAKD and DGKD can be summarized as follows: (I) Our method employs dual-teacher to teach different information, while TAKD and DGKD apply multiple teachers to teach the same information. (II) Although our method is inspired by DGKD, there is some difference between them. When distilling the student model, our method employs an orderly learning strategy, while DGKD utilizes multiple teachers to teach a student simultaneously. Crucially, our method can be seen as an extension of existing knowledge distillation methods.

\section{Approach}
\subsection{Background}
The key to knowledge distillation is to let a small network (student) imitate not only the output of a large network (teacher), but also true labels of datasets. Let ${l_{s}}$ and ${l_{t}}$ be the logits of the student and teacher, respectively. $\emph{T}$ is a temperature parameter to soften the output of the student and teacher. $y_{s}=softmax(l_{s}/T)$ and $y_{t}=softmax(l_{t}/T)$ are the soften outputs of the student and teacher, respectively. To encourage the student to mimic the output of the teacher, a KL-divergence
loss $L_{KD}$ can be minimized as follows:
\begin{equation}
L_{KD} = T^{2}KL(y_{s}, y_{t})
\label{Eq1}
\end{equation}
To minimize the gap between the output of the student model softmax($l_{s}$) and true labels $l$ of datasets, the cross-entropy loss $L_{CE}$ can be penalized as follows:
\begin{equation}
L_{CE} = F(softmax(l_{s}), l)
\label{Eq2}
\end{equation}
Finally, the overall loss function can be denoted by adding a balance factor $\alpha$ as follows:
\begin{equation}
L = (1 - \alpha)L_{CE} + \alpha L_{KD}
\label{Eq3}
\end{equation}

\subsection{ODKD Framework}
\begin{figure}[!htbp]
	\setlength{\belowcaptionskip}{-0.4cm}
	\begin{minipage}[b]{1.0\linewidth}
		\centering
		\centerline{\includegraphics[scale=0.20]{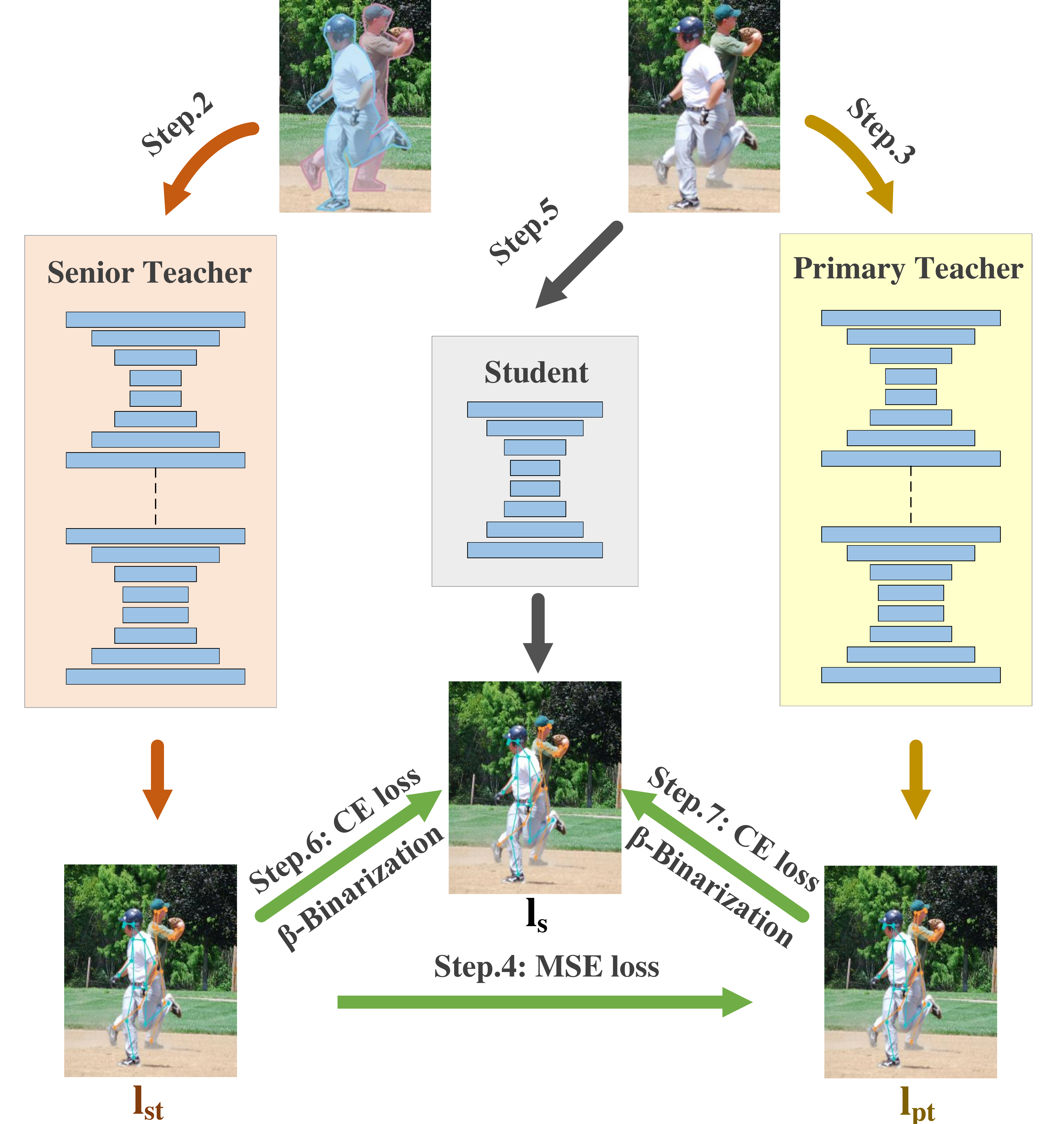}}
	\end{minipage}
	\caption{Illustration of our method. We adopt two loss functions: MSE loss and cross-entropy loss. $\beta$-Binarization denotes the binarization operation with a factor $\beta$.}
	\label{fig:network}
\end{figure}

As illustrated in Figure~\ref{fig:network}, our proposed orderly dual-teacher knowledge distillation (ODKD) framework consists of two large teacher networks with different capabilities: primary teacher (PT) and senior teacher (ST). The final target is to transfer knowledge of PT and ST to a lightweight student network. We adopt a binarization operation and two different loss functions. In this section, we firstly elaborate on the implementation process of an orderly dual-teacher. Secondly, we describe the details of a binarization operation. Finally, we give loss functions used in the overall process.

\textbf{Orderly dual-teacher.}
There are three differences between PT and ST. The first is the input of networks. The input of PT is a three-channel RGB image, while the input of ST is generated by concatenating a three-channel RGB image and one-channel human segmentation mask. The second is slightly different network structures. PT is a common human pose estimation model, such as SimpleBaseline~\cite{xiao2018simple}, while ST is the variant of PT where we add an $1\times1$ convolution at the head of ST to transform the number of channels from 4 to 3. The third is different capabilities. PT is used to teach keypoints information, but ST is used to transfer segmentation and keypoints information. 
There is some difference between a lightweight student and PT. For example, a student model employs fewer channels or stages than PT, which leads to a large performance gap between them. With the guidance of the human segmentation mask, ST is more powerful than PT, which further widens the gap between a student and ST. To solve the problem, as shown in Step 4 of Figure~\ref{fig:network}, we add a path where PT serves as a teacher assistant to bridge the gap between a student and ST. During training, we firstly pre-train ST by imitating true labels of datasets. Then PT is obtained by mimicking labels provided by datasets and ST. After these, an orderly learning strategy is utilized where a student learns from ST and PT successively, as shown in Steps 6 and 7 of Figure~\ref{fig:network}. At test time, only a student is employed.

\begin{figure}
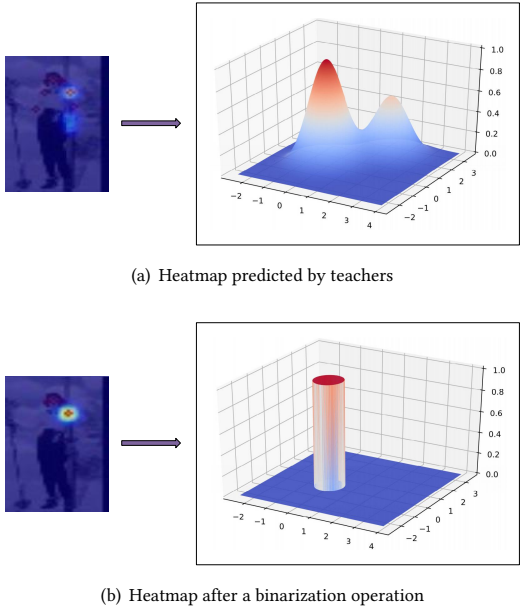

	\setlength{\belowcaptionskip}{-0.4cm}
	\subfigure[Heatmap predicted by teachers]{
		\includegraphics[scale=0.025]{ours4.pdf}
	}
	\subfigure[Heatmap after a binarization operation]{
		\includegraphics[scale=0.025]{ours3.pdf}
	}
	\caption{Illustration of a binarization operation. (a) Predicted Heatmap. (b) Heatmap representation after a binarization operation.}
	\label{heatmap}
\end{figure}
\textbf{Binarization operation.} The binarization operation aims to simplify the learning task and reduce noise in heatmaps generated by teachers. As shown in Figure~\ref{heatmap}(a), the heatmap predicted by teacher models almost exhibits a 2D Gaussian distribution structure where each pixel of the heatmap ranges from 0 to 1. It is very difficult for a lightweight student model to approximate the distribution. A straightforward solution is to convert the difficult task to a simple task. Moreover, multiple peaks appear in heatmaps generated by teachers and redundant peaks can be regarded as noise, which provides a worthless learning signal for a student. To solve the above problems, as shown in Figure~\ref{fig:network}, we employ a binarization operation during obtaining a lightweight student model, which can be denoted as follows with a threshold $\beta$:
\begin{equation}
\label{equ:E4}
C(y)=\left\{
\begin{aligned}
1 \quad if\ H(y) > \beta, \\
0 \quad if\ H(y) \leq \beta,
\end{aligned}
\right. 
\end{equation}
where $H(y)$ is the value of the heatmap at location $y$ and $C(y)$ is the class of pixel at location $y$. For the ground-truth heatmaps, we empirically set $\beta$ to 0.6. For the heatmaps from teachers, the threshold is chosen by the ablation experiments. After binarization, the result is illustrated in Figure~\ref{heatmap}(b). A student only needs to make a simple binary classification for each pixel and an appropriate threshold can eliminate redundant peaks. When training ST and PT, we do not employ a binarization operation as the number of layers of these two models is enough to approximate the data distribution.

\textbf{Loss functions.} Whether to use a binarization operation affects which loss function to use. When training ST and PT, a binarization operation is not employed, and therefore the loss function used is the conventional MSE loss, which can be denoted as follows:
\begin{equation}
\label{equ:E5}
L_{MSE} = \frac{1}{n} \sum_{i=1}^{n}(\hat{l}^{i} - l^{i})^{2},
\end{equation}
where $\hat{l}^{i}$ and $l^{i}$ specify the predicted heatmap and ground-truth heatmap for the $i$-th joint, respectively. In the process of obtaining a lightweight student model, a binarization operation is utilized, and the cross-entropy loss is penalized to minimize the gap between the predicted value $q=Sigmoid(l_{s})$ and the label value $p$:
\begin{equation}
\begin{split}
\label{equ:E6}
L_{C}&=-(plogq + (1-p)log(1-q))\\
&=\left\{
\begin{array}{lcl}
-logq & & if\ p = 1,\\
-log(1-q)          & & if\ p = 0. \\
\end{array} \right.
\end{split}
\end{equation}
\begin{algorithm}[htb]
	\setstretch{1.35} 
	\caption{: ODKD student training}
	\label{algorithm}
	\begin{algorithmic}
		\REQUIRE preprocessed image $x$ and human segmentation mask $m$, label $l$, pre-trained senior teacher $ST$, primary teacher $PT$, student $S$, the number of iterations $n$, the number of epochs $N$ ~~\\
		\ENSURE distilled student $S$.\\
		\FOR{$i=1$ to $N$}
		\FOR{$j=1$ to $n$}
		\STATE Step.1 concatenate $x$ and $m$ to obtain y;
		\STATE Step.2 feed $y$ to $ST$, to obtain the senior teacher logits $l_{st}$;\\
		\STATE Step.3 feed $x$ to $PT$, to obtain the primary teacher logits $l_{pt}$;\\
		\STATE Step.4 update $PT$ based on Eq.(\ref{equ:E8});
		\STATE Step.5 feed $x$ to $S$, to obtain the student logits $l_{s}$;\\
		\STATE Step.6 update $S$ based on Eq.(\ref{equ:E9});
		\STATE Step.7 update $S$ based on Eq.(\ref{equ:E10}).
		\ENDFOR
		\ENDFOR
	\end{algorithmic}
\end{algorithm}
Taking Eq.(\ref{Eq3}), (\ref{equ:E5}), and (\ref{equ:E6}) together, we can obtain the loss functions used in the overall process as follows:
\begin{equation}L_{ST} = L_{MSE}\end{equation}
\begin{equation}
\begin{split}
\label{equ:E8}
L_{PT} = L_{ST\rightarrow PT} &= (1 - \alpha_{0})L_{CE_{PT}} + \alpha_{0} L_{KD_{ST \rightarrow PT}}\\
&=(1 - \alpha_{0})L_{MSE_{PT}} + \alpha_{0} L_{MSE_{ST \rightarrow PT}},
\end{split}
\end{equation}
\begin{equation}
\begin{split}
\label{equ:E9}
L_{S_{1}} = L_{ST\rightarrow S} &= (1 - \alpha_{1})L_{CE_{S}} + \alpha_{1} L_{KD_{ST \rightarrow S}}\\
&=(1 - \alpha_{1})L_{C_{S}} + \alpha_{1} L_{C_{ST \rightarrow S}},
\end{split}
\end{equation}
\begin{equation}
\begin{split}
\label{equ:E10}
L_{S_{2}} = L_{PT\rightarrow S} &= (1 - \alpha_{2})L_{CE_{S}} + \alpha_{2} L_{KD_{PT \rightarrow S}}\\
&=(1 - \alpha_{2})L_{C_{S}} + \alpha_{2} L_{C_{PT \rightarrow S}},
\end{split}
\end{equation}
where the right arrow at the subscript indicates the teaching direction, and $\alpha_{0}$, $\alpha_{1}$ and $\alpha_{2}$ are balance factors. Here we set them to 0.5 as demonstrated in FPD~\cite{zhang2019fast}. Finally, the overall process can be summarized as Algorithm~\ref{algorithm}.
\section{Experiment}
\subsection{Experiment Setup}
\textbf{Dataset}. We employ two human pose estimation datasets, COCO and OCHuman. COCO dataset contains over 200\emph{K} images and 250\emph{K} person instances labeled with 17 keypoints. The images are extracted from real scenes. We train our models only on the train2017 set, equipped with 57\emph{K} images and 150\emph{K} person instances, and evaluate our method on the val2017 set and test-dev2017 set, consisting of 5\emph{K} images and 20\emph{K} images, respectively.

OCHuman dataset is also collected from real scenes. Different from COCO dataset, it is a more challenging dataset, where each human instance is heavily occluded by one or several others and the postures of the human bodies are more complex. The purpose of designing this dataset is to use general datasets such as COCO as a training set, test the robustness of models to occlusion using OCHuman. Therefore, this dataset has no training set, but only a validation set and test set. The validation set and test set have 4731 images and 8110 person instances in total.

\textbf{Evaluation Metric}. For the two datasets, we employ the same evaluation metric based on Object Keypoint Similarity (OKS). OKS can be calculated by:
\begin{equation}
OKS=\frac{\sum_{i}exp(-d_{i}^{2}/2s^{2}k_{i}^{2})\delta(v_{i}>0)}{\sum_{i}\delta(v_{i}>0)},
\end{equation}
where, $d_{i}$ is the Euclidean distance between each ground truth keypoint and corresponding detected keypoint, $v_{i}$ is the visibility flag of the ground truth, $s$ is the object scale, and $k_{i}$ is a per-keypoint constant that controls falloff. We report standard average precision and recall: AP (the mean of AP scores at OKS = 0.50, 0.55,$\ldots$, 0.90, 0.95), AP$^{50}$ (AP at OKS = 0.50), AP$^{75}$, AP$^{M}$ for medium objects, AP$^{L}$ for large objects and AR (the mean of AR scores at OKS = 0.50, 0.55,$\ldots$, 0.90, 0.95).

\textbf{Training}. Our models are implemented on two NVIDIA 2080Ti GPUs. We extend the human detection boxes to a fixed ratio, namely height: width = 4 : 3, and then crop the boxes from images. Finally, we resize the cropped images to a fixed size, for example, $256\times192$. In the experiments, we choose three combinations of ST, PT and student models. One combination is that SimpleBaseline-ResNet50~\cite{xiao2018simple} with the human segmentation mask, SimpleBaseline-ResNet50 and LPN-ResNet50~\cite{zhang2019simple} are chosen as ST, PT and student, respectively. One combination is an 8-stage hourglass~\cite{zhang2019fast} with the human segmentation mask, 8-stage hourglass and 4-stage hourglass. Another combination is HRNet-W32~\cite{sun2019deep} with the human segmentation mask, HRNet-W32 and HRNet-W16. We reproduce LPN without any tricks, including the attention mechanism, the iterative training strategy and $\beta$-Soft-Argmax. Other settings are the same as the original work.
For SimpleBaseline and Hourglass network, we adopt the same training strategies as in the original works. Following the structure of HRNet, we design a small network HRNet-W16 by reducing the number of basic channels to 16. The total epochs are set as 150, and the learning rate is dropped at the 120th and 140th epochs.

\textbf{Testing}. The top-down pipeline is adopted that first locates the human body by the person detectors and then applies the pose estimation models to acquire the detection results. 
For a fair comparison, we adopt the same person detectors provided by HRNet~\cite{sun2019deep} both for COCO validation and test-dev set. The human detection AP is 56.4 and 60.9 respectively. For OCHuman test set, we use ground-truth detection boxes. Following the common practice\cite{chen2018cascaded, xiao2018simple, sun2019deep, newell2016stacked}, we compute the heatmap by averaging the heatmaps of the original and flipped images. The final keypoints are obtained by adjusting a quarter offset in the direction from the highest response to the second highest response.

\subsection{Component Ablation Studies}
In this subsection, we conduct the ablation experiments on the COCO validation set to verify the effectiveness of our proposed components. We choose SimpleBaseline as the teacher model and LPN as the student model. By default, the input size $256\times192$ and the ResNet-50 backbone are used because they are less computational.

\begin{table}[!ht]
	\caption{The ablation study on the architecture of ODKD on the COCO validation set. We adopt MSE Loss. $\rightarrow$ indicates the teaching direction.}
	\label{table:table1}
	\renewcommand\arraystretch{1} 
	\centering
	\setlength{\tabcolsep}{4mm}{
		\begin{tabular}{c|l|c}
			\toprule[1.5pt]
			Group &Method                & AP    \\
			\midrule[1.5pt]
			1 &(a) S (LPN, baseline)        & 64.5       \\
			\hline
			\multirow{2}{*}{2}  &(b) ST$\rightarrow$S      &  64.9        \\ 
			&(c) PT$\rightarrow$S      &  65.0         \\
			\hline
			\multirow{2}{*}[-6px]{3}&(d) ST, PT$\rightarrow$S      &  65.0         \\
			&(e) PT$\rightarrow$S, ST$\rightarrow$S    &  64.7\\
			&(f) ST$\rightarrow$S, PT$\rightarrow$S    &  65.0\\
			\hline
			\multirow{2}{*}[-10px]{4}&(g) ST$\rightarrow$PT$\rightarrow$S      &  64.9         \\
			&(h) ST$\rightarrow$PT.  ST, PT$\rightarrow$S & 64.9\\
			&(i) ST$\rightarrow$PT. PT$\rightarrow$S, ST$\rightarrow$S				  & 65.0       \\
			&(j) ST$\rightarrow$PT. ST$\rightarrow$S, PT$\rightarrow$S (ODKD)                 & \textbf{65.2}  \\
			\bottomrule[1.5pt]
	\end{tabular}}
\end{table}
\begin{table}[htbp]
	\caption{Comparisons on the binarization thresholds and GPU memory usage. The batch size is set as 24. GMU denotes GPU Memory Usage. Here LPN is chosen as our baseline.}
	\label{table:table2}
	\renewcommand\arraystretch{1}
	\centering
	\setlength{\tabcolsep}{1mm}{
		\begin{tabular}{c|c|ccccc}
			\toprule[1.5pt]
			Models & ODKD w/o binarization & \multicolumn{5}{c}{ODKD w/ binarization} \\
			\hline
			$\beta$     &  -     & 0.2   & 0.3   & 0.4   & 0.5 & 0.6\\
			\hline
			AP    & 65.2  & 65.7  & \textbf{65.9}  & 65.7  & 65.3  & 65.1 \\
			\hline
			GMU (MB)   & 9286  & \multicolumn{5}{c}{\textbf{6694}} \\
			\bottomrule[1.5pt]
	\end{tabular}}
\end{table}%
\textbf{The studies on distillation paths.} We explore the performance of different distillation paths on the COCO validation set. As shown in Table~\ref{table:table1},
we divide all the experiments into four groups. The first group is our baseline model, which achieves 64.5 AP. The second group belongs to single teacher knowledge distillation. The third group belongs to dual-teacher knowledge distillation, where there is no path between ST and PT. The fourth group is dual-teacher knowledge distillation, where PT serves as a teacher assistant to bridge the gap between ST and S. We firstly compare these methods within each group. In group 2, Method (b) obtains less performance improvement than (c), as the output of ST is more abstract than that of PT, which is a more difficult learning task for a student. In group 3, Method (d) is a situation where dual-teacher teaches a student simultaneously, while Method (e) and (f) belong to a circumstance where dual-teacher teaches a student in multiple steps. Method (d) and (f) achieve 65.0 AP, which is higher than (e). The reason behind this is that in Method (d) and (f), segmentation information can be firstly used to help restrict the location range of keypoints, and then keypoints information is utilized to optimize the location of keypoints, which produces more accurate heatmaps. While in Method (e), segmentation information is learned in the second step, which leads to inadequate utilization of prior information. In group 4, Method (h) employs dual-teacher to guide a student simultaneously, while Method (i) and (j) utilize an orderly dual-teacher knowledge distillation strategy. Method (i) and (j) receive more improvement than (h), which proves the effectiveness of an orderly learning strategy. Method (j) gets higher AP than (i), which further demonstrates the superiority of firstly learning segmentation information. Comparisons of methods in different groups are shown as follows. Compared with group 1, all of the other groups achieve performance improvements, which shows the advantage of knowledge distillation. Group 2 and 3 achieve similar performance. 
\begin{table*}[!ht]
	\caption{Comparisons on the COCO validation set. The result of 8-stage Hourglass is cited from~\cite{newell2016stacked}. 8-stage Hourglass$^{*}$, 4-stage Hourglass$^{*}$ and Lite-HRNet$^{*}$: models reproduced by ourselves using the COCO dataset.}
	\label{table:table3}
	\centering
	\setlength{\tabcolsep}{1.2mm}{
		\begin{tabular}{l|l|c|c|c|c|c|cccccc}
			\toprule[1.5pt]
			Method                & Backbone   & Pretrain   & Input size  & \#Params  & FLOPs & FPS               & AP       & $\rm AP^{50}$   & $\rm AP^{75}$  & $\rm AP^{M}$  & $\rm AP^{L}$  & AR     \\
			\midrule[1.5pt]
			\multicolumn{13}{c}{\emph{Large networks}} \\
			\hline
			8-stage Hourglass~\cite{newell2016stacked}    & 8-stage Hourglass  & N  & $256\times192$  & 25.1M  & 14.3G & -  & 66.9 & -  & -  & -  & -  & -    \\ 
			HRNet-W32~\cite{sun2019deep}   & HRNet-W32  & N  &  $256\times192$  & 28.5M  & 7.1G &140 & 73.4  & 89.5  & 80.7  & 70.2  & 80.1  & 78.9 \\
			8-stage Hourglass$^{*}$~\cite{zhang2019fast}    & 8-stage Hourglass  & N  & $256\times192$  & 25.6M  & 21.3G &49 & 73.7 & 89.3  & 80.7  & 70.4  & 80.4  & 79.2   \\ 
			CPN~\cite{chen2018cascaded}              & ResNet-50                & Y   &$256\times192$  & 27.0M  & 6.2G &77             & 69.4     & -  & -  & -  & -  & -  \\
			SimpleBaseline~\cite{xiao2018simple}   & ResNet-50  & Y  &  $256\times192$  & 34.0M  & 8.9G &187  & 70.4  & 88.6  & 78.3  & 67.1  & 77.2  & 76.3\\
			SimpleBaseline~\cite{xiao2018simple}   & ResNet-101  & Y  &  $256\times192$  & 53.0M  & 12.4G &155 & 71.4  & 89.3  & 79.3  & 68.1  & 78.1  & 77.1 \\
			SimpleBaseline~\cite{xiao2018simple}   & ResNet-152  & Y  &  $256\times192$  & 68.6M  & 15.7G &124 & 72.0  & 89.3  & 79.8  & 68.7  & 78.9  & 77.8 \\
			HRNet-W32~\cite{sun2019deep}   & HRNet-W32  & Y  &  $256\times192$  & 28.5M  & 7.1G &140 & 74.4  & 90.5  & 81.9  & 70.8  & 81.0  & 79.8 \\
			HRNet-W48~\cite{sun2019deep}   & HRNet-W48  & Y  &  $256\times192$  & 63.6M  & 14.6G &103 & 75.1  & 90.6  & 82.2  & 71.5  & 81.8  & 80.4 \\
			\hline
			\multicolumn{13}{c}{\emph{Small networks}} \\
			\hline
			LPN50~\cite{zhang2019simple}  & ResNet-50  & N  &  $256\times192$  & \textbf{2.7M}  & \textbf{1.1G} &\textbf{243}  & 64.5  & 86.3 & 71.8 & 61.1 & 71.1 & 70.7\\
			\rowcolor{mygray}
			+\textbf{ODKD} & ResNet-50  & N  &  $256\times192$  & \textbf{2.7M}  & \textbf{1.1G} &\textbf{243} & \textbf{65.9(+1.4) } & \textbf{86.9} & \textbf{73.1} & \textbf{62.5} & \textbf{72.8} & \textbf{72.0}\\
			\hline
			4-stage Hourglass$^{*}$~\cite{zhang2019fast}  & 4-stage Hourglass  & N  &  $256\times192$  & \textbf{3.3M}  & \textbf{3.0G} &\textbf{158}  & 68.3  & 87.1 & 75.4 & 65.3 & 74.3 & 74.1\\
			\rowcolor{mygray}
			+\textbf{ODKD}  & 4-stage Hourglass  & N  &  $256\times192$  &\textbf{3.3M} &\textbf{3.0G} &\textbf{158} & \textbf{69.3(+1.0)}  & \textbf{87.4} & \textbf{76.6} & \textbf{66.2} & \textbf{75.8} & \textbf{75.1}\\
			\hline
			HRNet-W16~\cite{sun2019deep}  & HRNet-W16  & N  & $256\times192$  & \textbf{7.5M} &\textbf{2.6G} &\textbf{163} & 68.4  & 88.3  &76.7  &65.2  &74.7  &74.7\\
			\rowcolor{mygray}
			+\textbf{ODKD}  & HRNet-W16  & N  & $256\times192$  &\textbf{7.5M}  &\textbf{2.6G} &\textbf{163}  & \textbf{71.7(+3.3)}  & \textbf{89.3}  &\textbf{79.1}  & \textbf{68.7}  &\textbf{78.0}  &\textbf{77.5}\\
			\hline
			Lite-HRNet$^{*}$~\cite{Yulitehrnet21}  & Lite-HRNet-18  & N  &  $256\times192$  & \textbf{1.1M}  & \textbf{0.2G} &-  & 61.4  & 85.6 & 68.6 & 58.7 & 67.4 & 68.2\\
			\rowcolor{mygray}
			+\textbf{ODKD} & Lite-HRNet-18  & N  &  $256\times192$  & \textbf{1.1M}  & \textbf{0.2G} &-  & \textbf{62.1(+0.7)}  & \textbf{85.7} & \textbf{69.6} & \textbf{59.2} & \textbf{67.9} & \textbf{68.6}\\
			Lite-HRNet$^{*}$~\cite{Yulitehrnet21}  & Lite-HRNet-30  & N  &  $256\times192$  & \textbf{1.8M}  & \textbf{0.3G} &-  & 64.2  & 86.8 & 71.8 & 61.6 & 69.8 & 70.7\\
			\rowcolor{mygray}
			+\textbf{ODKD} & Lite-HRNet-30  & N  &  $256\times192$  & \textbf{1.8M}  & \textbf{0.3G} &-  & \textbf{65.6(+1.4)}  & \textbf{87.4} & \textbf{73.3} & \textbf{63.1} & \textbf{71.3} & \textbf{72.0}\\
			\bottomrule[1.5pt]
	\end{tabular}}
\end{table*}
By adding a path on the basis of group 3, group 4 can be obtained, which receives a better performance than the original, which demonstrates that it is necessary to introduce PT as a teacher assistant to narrow the gap between ST and S. Compared to Method (c), Method (j) obtains 0.2 AP improvement, which is not obvious. The main reason is the insufficient learning ability of a student, rather than an orderly dual-teacher learning strategy. Among all methods, Method (j) achieves the best 65.2 AP. Therefore, this setting is selected finally.

\begin{figure}[!htbp]
	\setlength{\belowcaptionskip}{-0.4cm}
	\begin{minipage}[b]{1.0\linewidth}
		\centering
		\centerline{\includegraphics[scale=0.55]{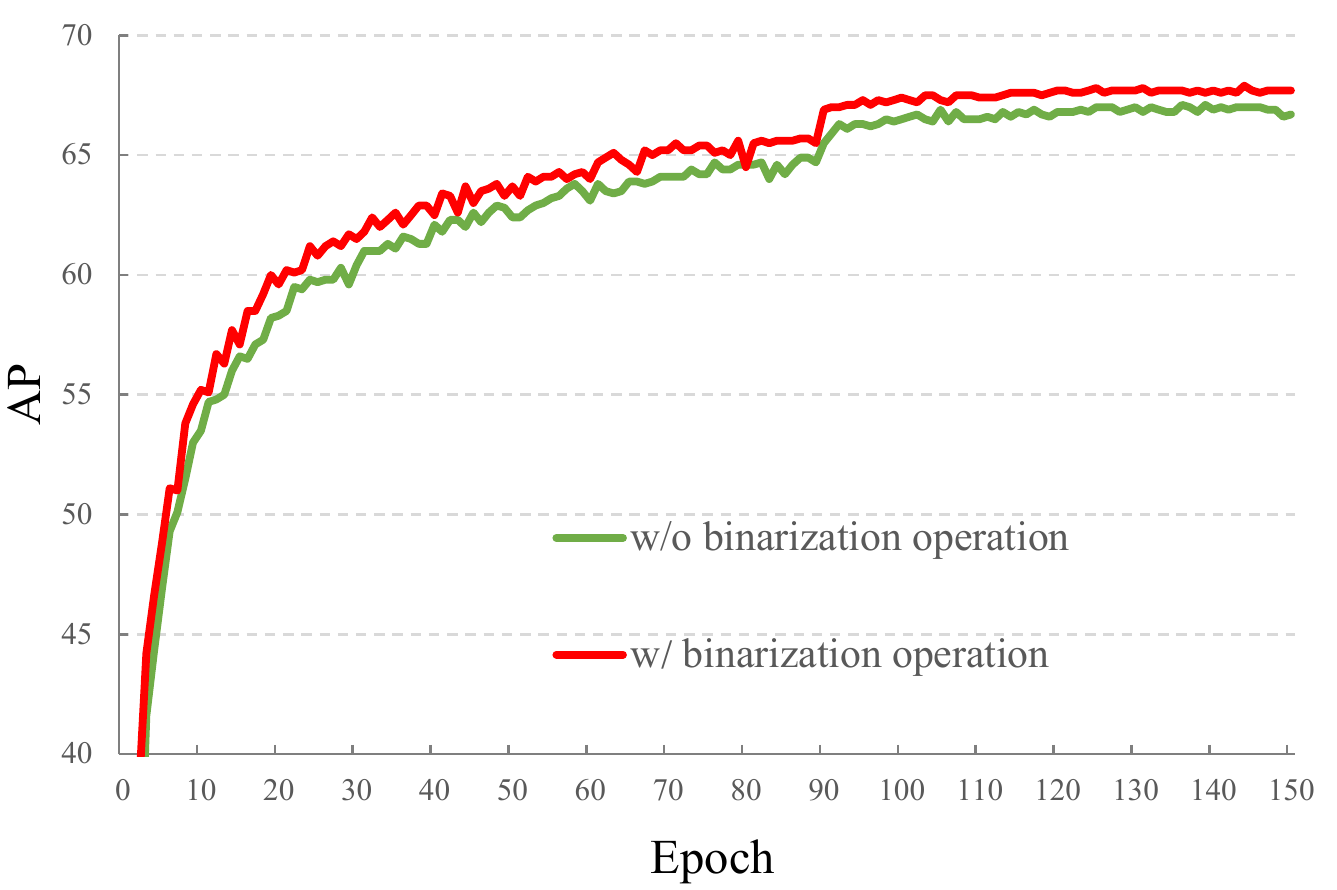}}
	\end{minipage}
	\caption{Illustration of performance with or without a binarization operation.}
	\label{fig:binary}
\end{figure}

\textbf{The binarization thresholds of heatmaps from teachers.}
We compare the effects of the heatmaps with different binarization thresholds on the COCO validation set. As shown in Table~\ref{table:table2}, with increasing $\beta$, the performance first increases and then decreases. The binarization thresholds represent the degree of reducing noise. The larger it is, the stronger the degree of reducing noise is. When $\beta$ is set to 0.3, lots of noise is removed and a model achieves the best 65.9 AP. When it is large enough such as 0.6, part of the learning signal in heatmaps will be eliminated, which leads to model degradation.  Finally,  $\beta$ is set as 0.3. A binarization operation can reduce GPU memory usage, which is friendly to resource-limited devices. Moreover, a binarization operation can effectively improve the model performance as illustrated in Figure~\ref{fig:binary}.

\begin{figure*}[!htbp]
	\begin{minipage}[b]{1.0\linewidth}
		\centering
		\centerline{\includegraphics[scale=0.55]{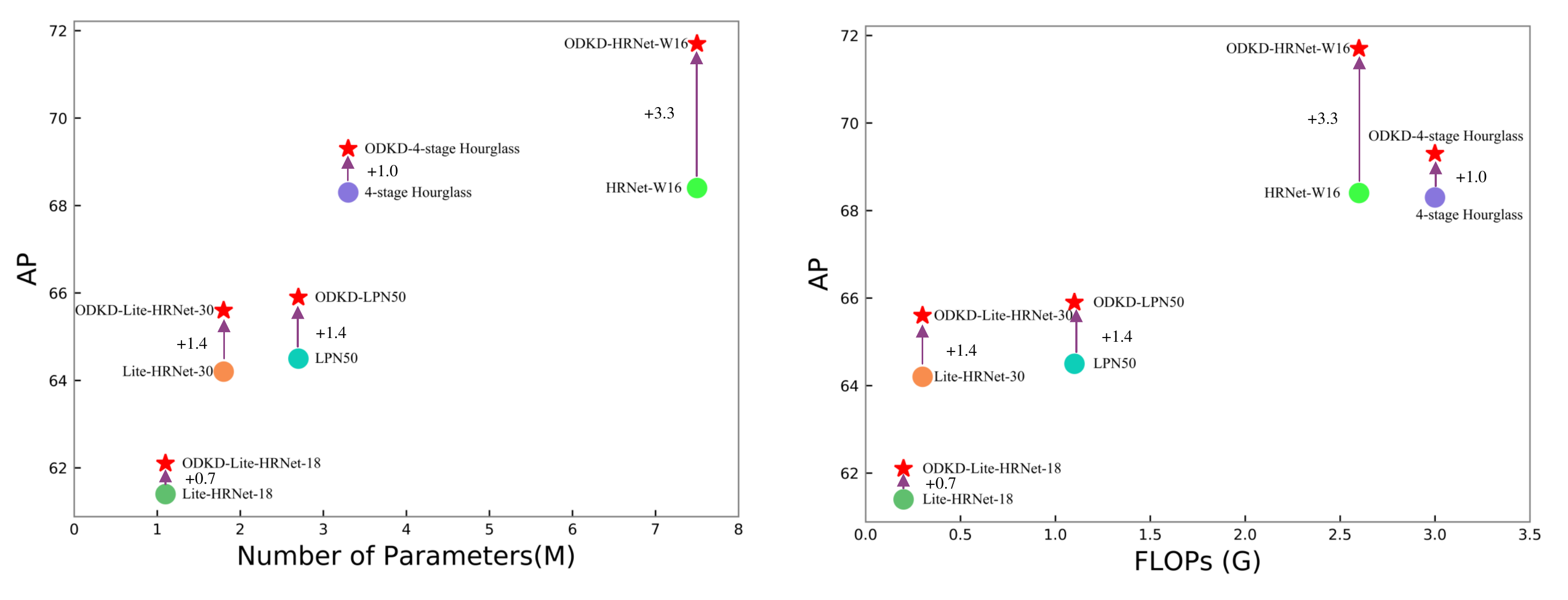}}
	\end{minipage}
	\caption{Illustration of the complexity and accuracy comparison on the COCO validation set.}
	\label{fig:param_flops}
\end{figure*}

\begin{table*}[!ht]
	\caption{Comparisons on the COCO test-dev set. 8-stage Hourglass$^{*}$, 4-stage Hourglass$^{*}$ and Lite-HRNet$^{*}$: models reproduced by ourselves using the COCO dataset.}
	\label{table:table4}
	\centering
	\setlength{\tabcolsep}{1.5mm}{
		\begin{tabular}{l|l|c|c|c|ccccccc}
			\toprule[1.5pt]
			Method                & Backbone         & Input size  & \#Params  & FLOPs     & AP       & $\rm AP^{50}$   & $\rm AP^{75}$  & $\rm AP^{M}$  & $\rm AP^{L}$  & AR    \\
			\midrule[1.5pt]
			\multicolumn{11}{c}{\emph{Large networks}} \\
			\hline
			OpenPose~\cite{cao2017realtime}              & -                & -  & -  &  -                & 61.8     & 84.9            & 67.5           & 57.1          & 68.2          & 66.5  \\
			Associative Embedding~\cite{newell2016associative} & -                & -  & -  & -              & 65.5     & 86.8            & 72.3           & 60.6          & 72.6          & 70.2  \\
			PersonLab\cite{papandreou2018personlab} & -                & -   & -  & -              & 68.7      & 89.0            & 75.4   & 64.1     & 75.5      & 75.4      \\
			HigherHRNet\cite{cheng2020higherhrnet} (multi-scale test)  & HRNet-W48  & $640\times640$  & 63.8M  & 154.3G   & 70.5  & 89.3  & 77.2  & 66.6  & 75.8 & 74.9 \\
			\hline
			Mask-RCNN\cite{he2017r}       & ResNet-50-FPN    & -   & -   & -  & 63.1  & 87.3  & 68.7  & 57.8  & 71.4  & -  \\
			G-RMI\cite{papandreou2017towards}         & ResNet-101  & $353\times257$   & 42.6M  & 57.0G  & 64.9  & 85.5  & 71.3  & 62.3  & 70.0  & 69.7  \\
			CPN~\cite{chen2018cascaded}                   & ResNet-50 & $256\times192$  & 27.0M  & 6.2G  & 68.6     & 89.5           & 76.6           & 65.6         & 74.2          & 75.6  \\
			RMPE~\cite{fang2017rmpe}      & PyraNet & $320\times256$  & 28.1M  & 26.7G    & 72.3  & 89.2  & 79.1  & 68.0  & 78.6  & -  \\
			SimpleBaseline~\cite{xiao2018simple}   & ResNet-50  &  $256\times192$  & 34.0M  & 8.9G  &70.0   &90.9   &77.9   &66.8   &75.8   &75.6  \\
			HRNet-W32~\cite{sun2019deep}  & HRNet-W32  & $256\times192$  & 28.5M  & 7.1G  &73.5   &92.2   &81.9   &70.2   &79.2   &79.0   \\
			HRNet-W48~\cite{sun2019deep}  & HRNet-W48  & $256\times192$  & 63.6M  & 14.6G  &74.2   &92.4   &82.4   &70.9   &79.7   &79.5   \\
			8-stage Hourglass$^{*}$~\cite{zhang2019fast}    & 8-stage Hourglass  & $256\times192$  & 25.6M  & 21.3G  & 73.2 & 91.3  & 81.1  & 70.2  & 79.0  & 78.7   \\ 
			\hline
			\multicolumn{11}{c}{\emph{Small networks}} \\
			\hline
			LPN50~\cite{zhang2019simple}  & ResNet-50  &  $256\times192$  & \textbf{2.7M}  & \textbf{1.1G}  & 64.2  & 88.6 & 71.2 & 61.0 & 69.8 & 70.1 \\
			\rowcolor{mygray}
			+\textbf{ODKD} & ResNet-50  &  $256\times192$  & \textbf{2.7M}  & \textbf{1.1G}  & \textbf{65.5(+1.3)}  & \textbf{89.2} & \textbf{72.8} & \textbf{62.4} & \textbf{71.1} & \textbf{71.4} \\
			\hline 
			4-stage Hourglass$^{*}$~\cite{zhang2019fast}  & 4-stage Hourglass   &  $256\times192$  &\textbf{3.3M}  &\textbf{3.0G}  & 67.8  & 89.1 & 75.4 & 64.9 & 73.2 & 73.4 \\
			\rowcolor{mygray}
			+\textbf{ODKD}  & 4-stage Hourglass    &  $256\times192$  &\textbf{3.3M} &\textbf{3.0G}  & \textbf{69.1(+1.3)}  &\textbf{89.9} &\textbf{76.7} &\textbf{66.0} &\textbf{74.7} &\textbf{74.7} \\
			\hline
			HRNet-W16~\cite{sun2019deep}  & HRNet-W16    & $256\times192$  & \textbf{7.5M} &\textbf{2.6G}  & 67.6  &90.2  &76.1  &64.7  &73.0  &73.7 \\
			\rowcolor{mygray}
			+\textbf{ODKD}  & HRNet-W16  & $256\times192$  &\textbf{7.5M}  &\textbf{2.6G}  & \textbf{71.0(+3.4)}  & \textbf{91.1}  &\textbf{79.5} &\textbf{67.9}  &\textbf{76.6}  &\textbf{76.6} \\
			\hline
			Lite-HRNet$^{*}$~\cite{Yulitehrnet21}  & Lite-HRNet-18  &  $256\times192$  & \textbf{1.1M}  & \textbf{0.2G}  & 61.1  & 87.6 & 68.0 & 58.8 & 65.9 & 68.0\\
			\rowcolor{mygray}
			+\textbf{ODKD} & Lite-HRNet-18  &  $256\times192$  & \textbf{1.1M}  & \textbf{0.2G}  & \textbf{61.8(+0.7)}  & \textbf{87.9} & \textbf{69.0} & \textbf{59.3} & \textbf{66.6} & \textbf{68.1}\\
			Lite-HRNet$^{*}$~\cite{Yulitehrnet21}  & Lite-HRNet-30  &  $256\times192$  & \textbf{1.8M}  & \textbf{0.3G}  & 63.9  & 88.9 & 71.6 & 61.4 & 68.8 & 70.3\\
			\rowcolor{mygray}
			+\textbf{ODKD} & Lite-HRNet-30  &  $256\times192$  & \textbf{1.8M}  & \textbf{0.3G}  & \textbf{65.2(+1.3)}  & \textbf{89.6} & \textbf{73.2} & \textbf{62.8} & \textbf{70.1} & \textbf{71.3}\\
			\bottomrule[1.5pt]
	\end{tabular}}
\end{table*}
\begin{figure*}[!htbp]
	\begin{minipage}[b]{1.0\linewidth}
		\centering
		\centerline{\includegraphics[scale=0.038]{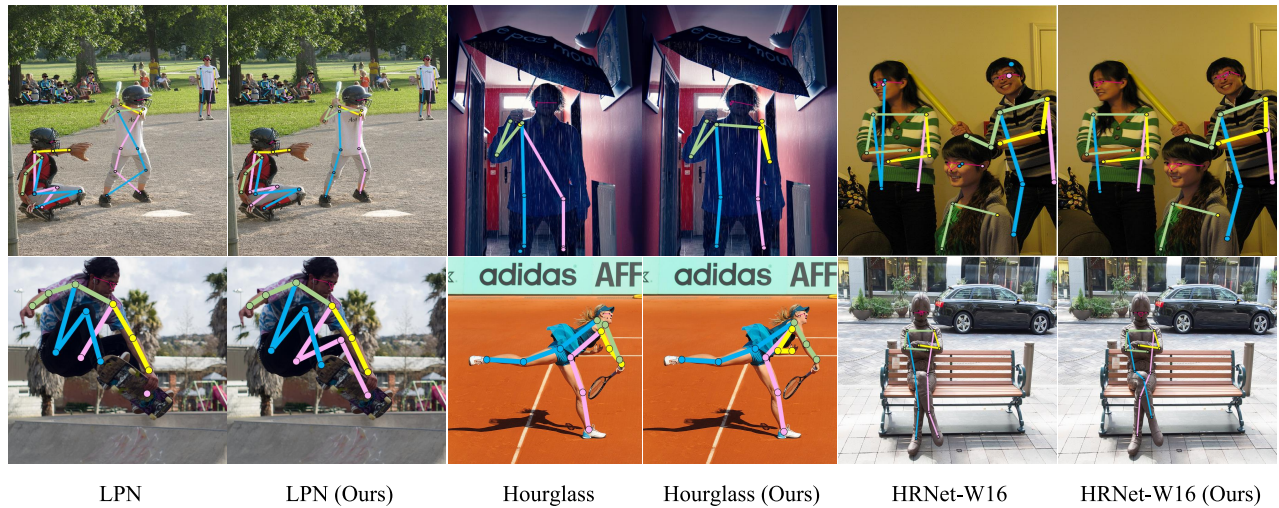}}
	\end{minipage}
	\caption{Qualitative results on COCO validation set.}
	\label{fig:visual}
\end{figure*}

\begin{table*}[!ht]
	\caption{Comparisons on the OCHuman test set. 8-stage Hourglass$^{*}$, 4-stage Hourglass$^{*}$ and Lite-HRNet$^{*}$: models reproduced by ourselves using the COCO dataset. There is no medium person instances, so $\rm AP^{M}$ denotes as -.}
	\label{table:table5}
	\renewcommand\arraystretch{1.1}
	\centering
	\setlength{\tabcolsep}{2.1mm}{
		\begin{tabular}{l|l|c|c|c|cccccc}
			\toprule[1.5pt]
			Method                & Backbone      & Input size  & \#Params  & FLOPs                & AP       & $\rm AP^{50}$   & $\rm AP^{75}$  & $\rm AP^{M}$  & $\rm AP^{L}$  & AR    \\
			\midrule[1.5pt]
			\multicolumn{11}{c}{\emph{Large networks}} \\
			\hline
			CPN~\cite{chen2018cascaded}              & ResNet-50                   &$256\times192$  & 27.0M  & 6.2G              & 52.8     & 75.8  & 55.2  & -  & 52.8  & 61.6   \\
			SimpleBaseline~\cite{xiao2018simple}   & ResNet-50    &  $256\times192$  & 34.0M  & 8.9G  & 55.8  & 73.9  & 61.3  & -  & 55.8  & 60.4 \\
			SimpleBaseline~\cite{xiao2018simple}   & ResNet-101    &  $256\times192$  & 53.0M  & 12.4G  & 60.0  & 76.0  & 66.1  & -  & 60.0  & 63.9 \\
			SimpleBaseline~\cite{xiao2018simple}   & ResNet-152    &  $256\times192$  & 68.6M  & 15.7G  & 61.9  & 77.0  & 67.8  & -  & 61.9  & 66.1 \\
			HRNet-W32~\cite{sun2019deep}   & HRNet-W32    &  $256\times192$  & 28.5M  & 7.1G  & 63.0  & 79.2  & 68.5  & -  & 63.0  & 66.9 \\
			HRNet-W48~\cite{sun2019deep}   & HRNet-W48   &  $256\times192$  & 63.6M  & 14.6G  & 64.6  & 79.6  & 70.7  & -  & 64.6  & 68.1 \\
			8-stage Hourglass$^{*}$~\cite{zhang2019fast}    & 8-stage Hourglass  & $256\times192$  & 25.6M  & 21.3G  &64.7  & 79.6  &69.9  & -  & 64.7  & 68.3   \\ 			
			\hline
			\multicolumn{11}{c}{\emph{Small networks}} \\
			\hline
			LPN50~\cite{zhang2019simple}  & ResNet-50    &  $256\times192$  & \textbf{2.7M}  & \textbf{1.1G} & 49.6  &\textbf{73.4}  & 54.2 & - & 49.6 & 54.9 \\
			\rowcolor{mygray}
			+\textbf{ODKD} & ResNet-50    &  $256\times192$  & \textbf{2.7M} & \textbf{1.1G} & \textbf{50.3(+0.7)}  & 72.1 &\textbf{54.6} &- &\textbf{50.3} &\textbf{55.9} \\
			\hline
			4-stage Hourglass$^{*}$~\cite{zhang2019fast}  & 4-stage Hourglass    &  $256\times192$  & \textbf{3.3M} &\textbf{3.0G}  & 56.3 & 75.0 & 62.6 & - & 56.3 & 60.9 \\
			\rowcolor{mygray}
			+\textbf{ODKD}  & 4-stage Hourglass    &  $256\times192$  &\textbf{3.3M} &\textbf{3.0G}  & \textbf{58.6(+2.3)}  & \textbf{77.0} &\textbf{64.6} & - &\textbf{58.6} &\textbf{63.0} \\
			\hline
			HRNet-W16~\cite{sun2019deep}  & HRNet-W16    & $256\times192$  & \textbf{7.5M} &\textbf{2.6G}  &56.9  &76.9  &63.3  &-  &56.9  &61.4 \\
			\rowcolor{mygray}
			+\textbf{ODKD}  & HRNet-W16  & $256\times192$  &\textbf{7.5M}  &\textbf{2.6G}  &\textbf{60.5(+3.6)}  & \textbf{79.3}  &\textbf{66.7}  &-  &\textbf{60.5}  &\textbf{64.9} \\
			\hline
			Lite-HRNet$^{*}$~\cite{Yulitehrnet21}  & Lite-HRNet-18  &  $256\times192$  & \textbf{1.1M}  & \textbf{0.2G}  & 48.8  & 73.3 & 53.7 & - & 48.8 & 54.2\\
			\rowcolor{mygray}
			+\textbf{ODKD} & Lite-HRNet-18  &  $256\times192$  & \textbf{1.1M}  & \textbf{0.2G}  & \textbf{50.5(+1.7)}  & \textbf{76.6} & \textbf{55.5} & - & \textbf{50.5} & \textbf{55.9}\\
			Lite-HRNet$^{*}$~\cite{Yulitehrnet21}  & Lite-HRNet-30  &  $256\times192$  & \textbf{1.8M}  & \textbf{0.3G}  & 53.0  & 77.0 & 58.2 & - & 53.0 & 57.9\\
			\rowcolor{mygray}
			+\textbf{ODKD} & Lite-HRNet-30  &  $256\times192$  & \textbf{1.8M}  & \textbf{0.3G}  & \textbf{54.1(+1.1)}  & \textbf{77.1} & \textbf{59.7} & - & \textbf{54.1} & \textbf{58.7}\\
			\bottomrule[1.5pt]
	\end{tabular}}
\end{table*}

\subsection{Experimental Results}
\textbf{Results on the COCO validation set.} We report the results of our method and other state-of-the-art methods in Table~\ref{table:table3}. (I) When ODKD is applied to lightweight models such as LPN50, 4-stage Hourglass and HRNet-W16, performance can be consistently improved, which demonstrates the superiority of the proposed ODKD. However, the improvements on different baseline models are discrepant. The reason for this is that the gaps between a student and teacher within each combination are different. The difference between LPN50 and SimpleBaseline~\cite{xiao2018simple} is that LPN50 employs a depth-wise separable convolution while SimpleBaseline uses an ordinary convolution. There are two differences between 4-stage Hourglass and 8-stage Hourglass. One is different numbers of stages, the other is different numbers of channels. The only difference between HRNet-W16 and HRNet-W32 is the number of channels. The model similarity between HRNet-W16 and HRNet-W32 is higher than other combinations, which yields higher performance improvement.
(II) Compared to 8-stage Hourglass and CPN~\cite{chen2018cascaded}, LPN50 and 4-stage Hourglass based on ODKD achieve comparable performance, while the parameters and calculation are much less than them. (III) HRNet-W16 trained with ODKD obtains better performance than SimpleBaseline with ResNet-50 and ResNet-101. And we can find that HRNet-W16 with 2.6GFLOPs runs slower than SimpleBaseline-ResNet50 with 8.9GFLOPs because there are lots of parallel convolutions in HRNet and PyTorch framework is not friendly to parallel convolutions. (IV) Although HRNet-W32 and HRNet-W48 achieve top accuracy, inference time is longer than our models. (V) Compared to the latest Lite-HRNet\footnote{Available from https://github.com/HRNet/Lite-HRNet. Because this paper and the corresponding code have just been published for few days, the experiments of adding ODKD to Lite-HRNet are not finished and the results are not displayed.}, our models based on ODKD achieve a better balance between accuracy and computational complexity, as shown in Figure~\ref{fig:param_flops}. We think that the proposed ODKD is a general framework and can be applied to Lite-HRNet to further improve the performance of the model. Some results generated by baseline models and our models are illustrated in Figure~\ref{fig:visual}. We can see that our models can work well in challenging situations.

\textbf{Results on the COCO test-dev set.} Table~\ref{table:table4} shows the results of our method and other state-of-the-art methods. Our proposed ODKD can promote LPN50 and 4-stage Hourglass by 1.3 AP. For HRNet-W16, the improvement is 3.4 AP. Compared to the bottom-up approaches, our models achieve acceptable results with fewer parameters and FLOPs. HRNet-W16 based on ODKD is significantly better than CPN~\cite{chen2018cascaded} and SimpleBaseline~\cite{xiao2018simple}. Compared to HRNet-W32 and HRNet-W48, HRNet-W16 trained with ODKD obtains close performance without pre-training. Compared to Lite-HRNet, our models achieve better performance with similar parameters and smaller input size. Moreover, HRNet-W16 equipped with ODKD sets a new state-of-the-art for lightweight human pose estimation. 

\textbf{Results on the OCHuman test set.}  We show the results of our methods and other state-of-the-art methods in Table~\ref{table:table5}. We use the models trained on COCO dataset to evaluate on OCHuman test set. As OCHuman dataset is more challenging than COCO dataset, the performance of all methods drops sharply. Different baseline models can benefit from ODKD, which proves the effectiveness and extensibility of ODKD. Moreover, our models based on ODKD use fewer parameters achieve comparable accuracy with large models, which is more suitable for mobile devices.

\section{CONCLUSION}
In this paper, we propose an orderly dual-teacher knowledge distillation (ODKD) framework for human pose estimation. Dual-teacher is introduced, where one is used to teach keypoints information to a student, the other is utilized to transfer segmentation and keypoints information. To improve the learning ability of a student, an orderly learning strategy is adopted where a student learns from dual-teacher successively. Moreover, a binarization operation is employed to stimulate more efficient learning of the a student and reduce noise in heatmaps generated by teachers. The experimental results on COCO and OCHuman keypoints dataset demonstrate the effectiveness and extensibility of our proposed ODKD.


\bibliographystyle{ACM-Reference-Format}
\bibliography{reference}

\appendix

\end{document}